# Evaluating Large Language Models in Crisis Detection: A Real-World Benchmark from Psychological Support Hotlines

Guifeng Deng, Shuying Rao, Tianyu Lin, Anlu Dai, Pan Wang, Junyi Xie, Yue Pan, Ke Zhao, Dongwu Xu, Zhengdong Cheng, Tao Li, Haiteng Jiang

***Abstract*—Psychological support hotlines serve as critical lifelines for crisis intervention but encounter significant challenges due to rising demand and limited resources. Large language models (LLMs) offer potential support in crisis assessments, yet their effectiveness in emotionally sensitive, real-world clinical settings remains underexplored. We introduce PsyCrisisBench, a comprehensive benchmark of 540 annotated transcripts from the Hangzhou Psychological Assistance Hotline, assessing four key tasks: mood status recognition, suicidal ideation detection, suicide plan identification, and risk assessment. 64 LLMs across 15 model families—including closed-source (e.g., GPT, Claude, Gemini) and open-source (e.g., Llama, Qwen, DeepSeek)—were evaluated using zero-shot, few-shot, and fine-tuning paradigms. LLMs showed strong results in suicidal ideation detection (F1=0.880), suicide plan identification (F1=0.779), and risk assessment (F1=0.907), with notable gains from few-shot prompting and fine-tuning. Compared to trained human operators, LLMs achieved comparable or superior performance on suicide plan identification and risk assessment, while humans retained advantages on mood status recognition and suicidal ideation detection. Mood status recognition remained challenging (max F1=0.709), likely due to missing vocal cues and semantic ambiguity. Notably, a fine-tuned 1.5B-parameter model (Qwen2.5-1.5B) outperformed larger models on mood and suicidal ideation tasks. LLMs demonstrate performance broadly comparable to trained human operators in text-based crisis assessment, with complementary strengths across task types. PsyCrisisBench provides a robust, real-world evaluation framework to guide future model development and ethical deployment in clinical mental health.**

## I. Introduction

Mental health conditions, including suicidal ideation and behaviors, remain a significant global public health challenge [1]. Psychological support hotlines offer a critical front line in crisis intervention [2], [3] yet the widening gap between growing demand and limited supply of trained human counselors raises concerns about system capacity and equity [4]. The rapid development of LLMs—notably GPT-4 [5], [6], Claude [7], Gemini [8], and ERNIE [9] —has introduced AI systems with remarkable language comprehension capabilities [10], [11]. In parallel, the emergence of open-source LLMs such as LLaMA [12], DeepSeek [13] and Qwen [14] has further broadened the landscape. These advances have sparked growing interest in leveraging LLMs to automatically assess suicide risk in hotline conversations, which may enable more efficient resource allocation and enhance the overall effectiveness of suicide prevention efforts.

LLMs have demonstrated strong performance across various mental health natural language processing (NLP) tasks [15], [16], such as clinical question answering [17], diagnosis [18], and summarization [19]. In the domain of psychological crisis hotlines, recent studies have applied deep learning and LLMs to predict future suicidal behavior from call recordings [20], [21], or employed BERT-based models to detect real-time risk in crisis chat [22]. However, these efforts focus on model development for specific tasks rather than systematic benchmarking of LLM capabilities. In parallel, benchmarking frameworks in the broader medical domain—such as PubMedQA [23], MultiMedQA [24], CORAL [25], PsyEval

This work was supported in part by STI2030-Major Projects (2022ZD0212400, 2021ZD0200404) National Natural Science Foundation of China (82371453), Key R&D Program of Zhejiang (2024C03006, 2024C04024, 2024ZY01010), Fundamental Research Funds for the Central Universities (2025ZFJH01-01) and the Construction Fund of Key Medical Disciplines of Hangzhou (2025HZGF10) *(Corresponding authors: Tao Li; Haiteng Jiang)*. Guifeng Deng and Shuying Rao are co-first authors.

Guifeng Deng and Shuyin Rao are with the College of Biomedical Engineering & Instrument Science, Zhejiang University, Hangzhou 310058, China, and also with the Affiliated Mental Health Center & Hangzhou Seventh People's Hospital, School of Brain Science and Brain Medicine, and Liangzhu Laboratory, Zhejiang University School of Medicine, Hangzhou 310058, China. (e-mail: 12415024@zju.edu.cn; 12215010@zju.edu.cn).

Tianyu Lin is with the Whiting School of Engineering, Johns Hopkins University, Maryland, 21218, USA. (e-mail: tlin67@jh.edu).

Anlu Dai, Pan Wang, Ke Zhao and Dongwu Xu are with the Department of Psychiatry and Mental Health, Wenzhou Medical University, Wenzhou 325035, Zhejiang Province, China. (e-mail: daianlu@wmu.edu.cn; wangpan@wmu.edu.cn; zhaoke@wmu.edu.cn; xdw@wmu.edu.cn).

Junyi Xie, Yue Pan, Haidong Song, Tao Li and Haiteng Jiang are with the Affiliated Mental Health Center & Hangzhou Seventh People's Hospital, School of Brain Science and Brain Medicine, and Liangzhu Laboratory, Zhejiang University School of Medicine, Hangzhou 310058, China. (e-mail: junyixie@zju.edu.cn; 22318470@zju.edu.cn; litaozjusc@zju.edu.cn; h.jiang@zju.edu.cn).

Zhengdong Cheng is with the College of Chemical and Biological Engineering, Zhejiang University, Hangzhou 310058, China. (e-mail: zcheng01@zju.edu.cn).

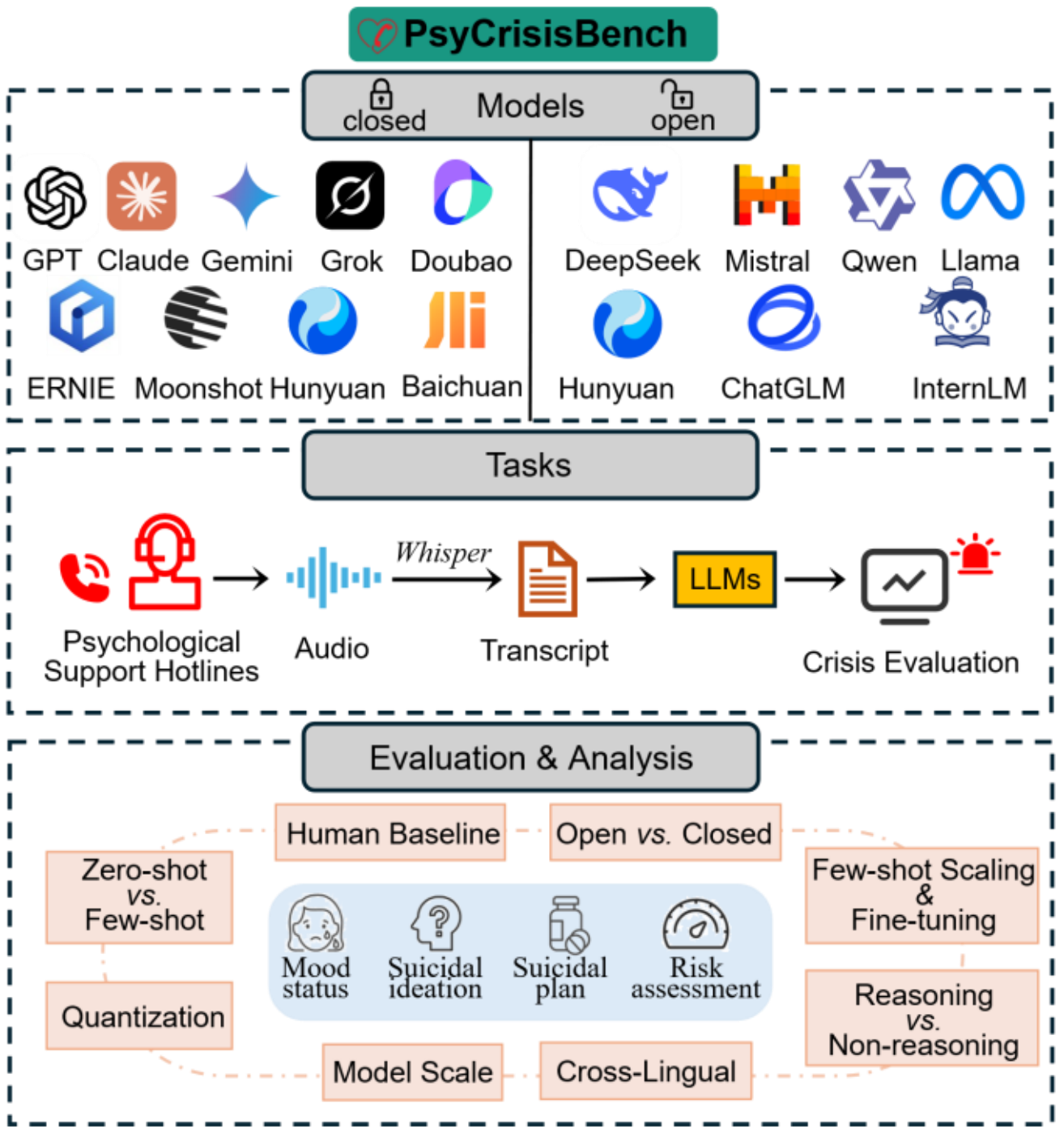

Fig. 1. PsyCrisisBench research framework overview. The top panel shows the 64 LLMs from 15 model families, including both open-source and closed-source models. The middle panel illustrates the crisis assessment workflow: psychological support hotline recordings are transcribed via Whisper and processed by LLMs for crisis evaluation. The bottom panel outlines the four core tasks and eight analytical dimensions: human baseline comparison, open- versus closed-source models, zero-shot versus few-shot prompting, few-shot scaling and fine-tuning effects, reasoning versus non-reasoning modes, quantization trade-offs, model scale effects, and cross-lingual generalizability.

[26], and PsychBench [27]—have established systematic evaluation protocols for LLMs in clinical contexts [28]. However, there is a lack of standardized evaluation for the complex, emotionally sensitive, and ethically nuanced nature of psychological crisis scenarios.

To address this need, we propose PsyCrisisBench, a comprehensive benchmarking framework grounded in real-world data from the Hangzhou Psychological Assistance Hotline at Hangzhou Seventh People's Hospital. The dataset includes 540 annotated hotline calls with demographic information, mood status, and suicide-related risk factors. PsyCrisisBench evaluates LLMs across four core crisis assessment tasks derived from the hotline's standardized clinical workflow: mood status recognition, suicidal ideation detection, suicide plan identification, and risk assessment. We systematically evaluate 64 models spanning 15 leading LLM families (Fig. 1), covering both closed-source and open-source models, across zero-shot, few-shot, and fine-tuned paradigms. Our goal is to bridge the current empirical evidence gap, offer a holistic view of model capabilities, and support the responsible integration of LLMs into psychological crisis response systems.

## II. Dataset

The data for PsyCrisisBench originates from the Hangzhou Psychological Assistance Hotline, which provides a free, 24-hour service nationwide. All hotline operators receive training in suicide risk assessment before handling calls. Operators collect basic demographic information from callers (e.g., gender, age, education level, marital status, occupation) as general statistical data. Additionally, during the consultation, the caller's mood status, suicide risk, and crisis severity are assessed, and structured annotations are completed after the call. This study received ethical approval from the Ethics Committee of the Seventh People's Hospital of Hangzhou. Callers are informed via a voice prompt before the call that the conversation will be recorded and anonymously documented for research and service improvement purposes.

Calls were collected from January 1, 2023, to December 31, 2023. From a total of 21,527 calls, the following were excluded: (1) invalid calls (duration less than 60 seconds, silent calls, or harassment calls); (2) calls seeking only information rather than psychological help; (3) for repeat callers, only the first call was retained. Based on the demographic data of high-risk calls (270 cases), non-high-risk calls matched for call time, gender, marital status, and call type were selected as a control group (270 cases), forming a dataset of 540 call audios. All call audios were transcribed into text using a locally deployed Whisper (whisper-large-v3-turbo) model [29]. The implementation of this transcription pipeline is publicly available at https://github.com/Deng-GuiFeng/multichannel-asr. To ensure privacy and de-identification, all transcripts were manually reviewed using a custom application to remove personally identifiable information, including specific addresses, names, phone numbers, and other unique identifiers. The transcripts underwent post-processing to normalize them into turn-by-turn dialogues (Fig. 2(a)). The demographic and call characteristics of PsyCrisisBench are summarized in Fig. 2(b), including distributions of gender, age, audio duration, and transcript word count. The average audio duration was 33 ± 25.2 minutes, and the average transcript word count was 6,489 ± 5,396 words per call. Each call was annotated with four binary labels derived from a standardized clinical assessment questionnaire (Table III): mood status (presence of depressive symptoms), suicidal ideation (expression of self-harm or suicidal thoughts), suicide plan (description of specific plan elements such as time, method, or means), and risk level (comprehensive assessment of intervention urgency). The class distribution for these labels is shown in Fig. 2(c). The detailed annotation protocol and judging criteria are provided in Appendix A.

Furthermore, to prevent data leakage and ensure fair model evaluation, an independent dataset was compiled for few-shot prompting and fine-tuning experiments. This auxiliary dataset consists of 520 calls selected from 16,573 calls made between January 1, 2022, and December 31, 2022, processed using the same methodology and annotation criteria as PsyCrisisBench. The detailed statistics of this auxiliary dataset are shown in Fig. 9.

## III. Methods

### A. Models Selection

This study included 64 LLMs spanning 15 major model families to comprehensively evaluate their performance differences in the context of psychological crisis hotlines, as illustrated in Fig. 1. Model selection followed these principles:

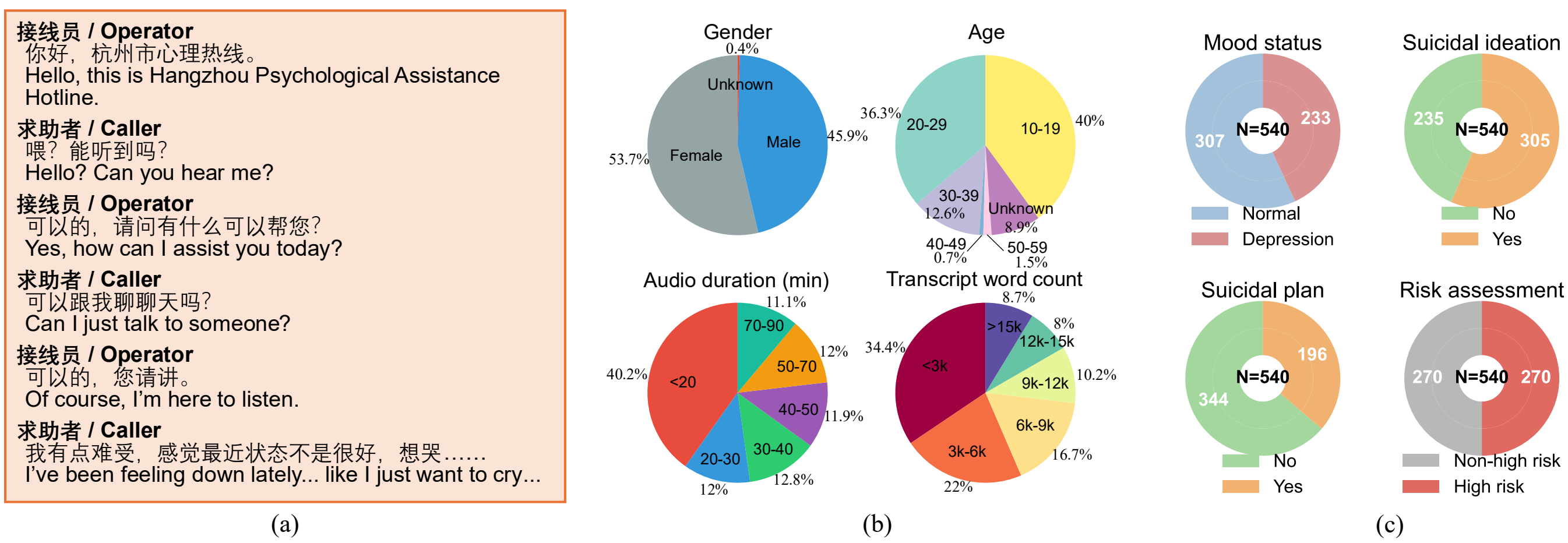

Fig. 2. Data Example and Statistics of the PsyCrisisBench Dataset. (a) An anonymized call excerpt (in Chinese with English translation) from the PsyCrisisBench dataset. (b) Pie charts showing the demographic and call characteristics of the 540 calls, including distribution of gender, age, audio duration in minutes, and transcript word count. (c) Class distribution for the four binary classification tasks: mood status (Depression: 233, Normal: 307), suicidal ideation (Yes: 305, No: 235), suicide plan (Yes: 196, No: 344), and risk assessment (High risk: 270, Non-high risk: 270).

first, leading model vendors with significant representation in the Chinese language domain were prioritized; subsequently, the current flagship models from each vendor were selected for evaluation. The chosen models cover a broad spectrum of influential LLMs in the current Chinese language context.

Among the 15 model series, closed-source models include GPT, Claude, Gemini, Grok, Doubao, ERNIE, Moonshot, and Baichuan. Open-source models cover mainstream series such as DeepSeek, Mistral, Qwen, Llama, and ChatGLM. Additionally, the Hunyuan series includes both open-source and closed-source models. In terms of distribution, of the 64 models, 26 are closed-source and 38 are open-source. Model parameter sizes range from 0.5 billion to trillion-scale (e.g., GPT-4, Claude-3, Gemini-2 series, whose exact parameter counts are undisclosed but estimated to exceed 1 trillion).

Based on their inference mechanisms, models were further categorized into reasoning and non-reasoning models. Reasoning models generate internal deliberation tokens before producing the final answer, whereas non-reasoning models (typically chat-oriented) provide direct responses without explicit intermediate reasoning steps. Reasoning models in this study include GPT-o1, GPT-o3, GPT-o3-mini, GPT-o4-mini, Claude-3.7-Sonnet, Gemini-2.5-Pro, Grok-3-mini, Doubao1.5-thinking-Pro, Hunyuan-T1, ERNIE-X1-Turbo, the DeepSeek-R1 series and its distilled variants, QwQ-32B, and GLM-Z1-32B-0414. Hybrid-reasoning models capable of switching between reasoning and non-reasoning modes include Gemini-2.5-Flash (evaluated in non-reasoning mode only) and the Qwen3 series (evaluated in both modes). All remaining models were classified as non-reasoning models.

All 64 models participated in the zero-shot evaluation. Building on this, considering differences in model support for context length, 38 models with context window lengths exceeding 64K were further subjected to static few-shot evaluation [30]. To further investigate the impact of the number of few-shot examples on model performance, the Qwen-2.5-7B-Instruct-1M model, which has long-context capabilities, was selected for dynamic few-shot evaluation experiments. Moreover, to address the common scenario where LLMs require fine-tuning with a small amount of domain-specific data to enhance task performance in practical applications, the Qwen2.5-1.5B-Instruct model [31] was chosen for fine-tuning experiments using the 2022 auxiliary dataset, to study the effect of fine-tuning on task performance. In the model deployment phase, to further reduce computational resource consumption and improve inference efficiency, AWQ quantized versions [32] of open-source models Qwen3 series (4B/8B/14B/32B), and QwQ-32B were included for comparative experiments against their full-precision counterparts. All models used in this study were officially released versions, with all open-source models being instruction-tuned variants. No task-specific modifications were made to the models except for the fine-tuning experiments, ensuring the fairness and reproducibility of the benchmark evaluation.

### B. Evaluation Protocols

The input for all models was the dialogue text from PsyCrisisBench. Models completed the binary classification for the four dimensions based on this text. The system prompt required models to output responses in a structured JSON format [33]; if this instruction was not followed, the response was marked as incorrect even if the answer was substantively correct.

Considering the inherent randomness in LLM outputs, repeated evaluations were conducted to ensure stability and reliability. For standard experiments (zero-shot, static few-shot, fine-tuning, quantization, and cross-lingual experiments), each model’s evaluation was repeated three times on the full PsyCrisisBench dataset. For dynamic few-shot experiments, each shot condition was repeated 10 times with different randomly sampled examples.

F1-score was chosen as the primary evaluation metric. The F1-score effectively balances precision and recall, which is particularly important for crisis assessment scenarios where both false negatives (missing high-risk individuals) and false positives (unnecessary interventions) carry significant consequences. To quantify uncertainty in reported F1-scores, 95% confidence intervals (CIs) were computed using the bootstrap method. Specifically, predictions from all repeated

TABLE I

SYSTEM PROMPTS USED FOR ZERO-SHOT, FEW-SHOT AND FINE-TUNING.

| Original Chinese | Translated English |
| --- | --- |
| 下面是杭州市心理援助热线的一个来电，请你通过对通话内容进行逐步分析，从”情绪状态”、”自杀意念”、”自杀计划”、”是否高危”四个角度评估来电求助者的抑郁状态和自杀危险性，其中情绪状态：抑郁或正常；自杀意念：有或无；自杀计划：有或无；以及，是否高危：高危或非高危。<br>若是存在严重心理危机，当下有比较紧迫的自杀风险，需要立马进行干预则是高危来电，若是通过倾诉沟通，心理问题得以疏导，情绪恢复稳定或建议就诊来获得更长期的帮助则是非高危来电。<br>需要注意的是，通话内容是使用语音识别模型转写的文本，识别结果存在许多错误，并且存在许多语气词，需要你在分析的过程中有效辨别出错误文本，忽略错误语义或者通过上下文重建正确语义。<br>最终判断要以 JSON 的形式输出，输出的 JSON 需遵守以下的格式：<br>```json<br>{<br>“情绪状态”: “<抑郁/正常>“,<br>“自杀意念”: “<有/无>“,<br>“自杀计划”: “<有/无>“,<br>“是否高危”: “<高危/非高危>“<br>}<br>```<br>下面我将给出 4 个样例供你学习<br># 样例 1 #<br># 样例 2 #<br># 样例 3 #<br># 样例 4 #<br>现在请你分析下面这个来电通话文本： | The following is a call transcript from the Hangzhou Psychological Assistance Hotline. Please analyze the conversation step-by-step to assess the caller's depressive state and suicide risk across four dimensions:<br>- Mood Status: Depression or Normal.<br>- Suicidal Ideation: Yes or No.<br>- Suicidal Plan: Yes or No.<br>- Risk Level: High risk or Non-high risk.<br>A call is classified as high risk if there is an immediate and severe psychological crisis requiring urgent intervention (e.g., imminent suicide risk). A call is classified as non-high risk if the caller's psychological distress can be alleviated through conversation (e.g., emotional stabilization achieved or referral to long-term professional help).<br>It's notable that the transcript is generated by a speech recognition model and may contain errors (e.g., misrecognized words, filler words). You must:<br>- Identify and disregard nonsensical phrases.<br>- Reconstruct accurate semantics from context when errors occur.<br>Return a JSON object strictly following this structure:<br>```json<br>{<br>“Mood status”: “< Depression/Normal>“,<br>“Suicidal ideation”: “<Yes/No>“,<br>“Suicidal plan”: “<Yes/No>“,<br>“Risk level”: “<High risk/Non-high risk>“<br>}<br>```<br>Below are four sample examples for your reference:<br># Example 1 #<br># Example 2 #<br># Example 3 #<br># Example 4 #<br>Now, please analyze the following hotline transcript: |

runs were pooled (e.g., 540 × 3 = 1,620 samples for standard experiments, or 540 × 10 = 5,400 samples for dynamic few-shot experiments), then resampled with replacement 10,000 times. The F1-score was calculated for each bootstrap sample, and the 2.5th and 97.5th percentiles of the resulting distribution were taken as the lower and upper bounds of the 95% CI.

Welch's t-test [34] was used for statistical comparisons when evaluating performance differences between different models or model groups (e.g., open-source vs. closed-source models, reasoning vs. non-reasoning models). Welch's t-test was chosen because it does not assume equal variances between groups, making it more appropriate for comparing model performances that may exhibit heterogeneous variability. The significance threshold was set at $p < 0.05$, with $p < 0.01$ indicating high significance.

### C. Experimental Settings

To establish a human performance reference, two trained hotline operators independently evaluated the 540 transcripts in PsyCrisisBench, with each operator assessing 270 cases. The operators performed the same four binary classification tasks using only the anonymized text transcripts, without access to the original audio recordings or the ground-truth labels.

The system prompts used in this study are presented in Table I. For zero-shot and fine-tuning evaluations, all models received the main instruction block (excluding the red-font examples) and a call transcript without examples. For static few-shot evaluation, models were provided with four fixed examples (highlighted in red) from the 2022 auxiliary dataset, selected to ensure class balance and appropriate transcript length (Table IV in Appendix B). Dynamic few-shot evaluation followed the same structure but with varying numbers of examples (0, 2, 4, 6, and 8 shots), randomly sampled for each of 10 repeated trials per condition. All LLM evaluations used Chinese transcripts, while the cross-lingual experiment used English translations.

To investigate the effect of example quantity on model performance, dynamic few-shot experiments were conducted using the Qwen2.5-7B-Instruct-1M model, which supports extended context lengths. Unlike static few-shot where examples were fixed, examples were randomly sampled from the 2022 auxiliary dataset for each trial. Each shot condition (0, 2, 4, 6, and 8 shots) was repeated 10 times with different randomly sampled examples to capture the variance introduced by example selection.

The fine-tuning experiment was conducted on Qwen2.5-1.5B-Instruct using the 520 calls from the 2022 auxiliary dataset (Fig. 9). Training used Fused AdamW optimizer for 3 epochs with learning rate 1e-5 and bfloat16 precision. Per-GPU batch size was 1 across 2 GPUs with gradient accumulation steps of 8 (effective batch size 16). Training and validation sets were split 4:1, with the lowest validation loss checkpoint selected for inference.

For quantization analysis, AWQ (Activation-aware Weight Quantization) [32], a low-bit weight-only quantization method that identifies and protects salient weights based on activation distribution, was applied to Qwen3 (4B/8B/14B/32B) and QwQ-32B. The quantized models were compared against their full-precision counterparts under zero-shot evaluation with identical inference parameters. Average peak GPU memory usage was recorded during inference as a measure of computational resource requirements.

To evaluate the generalizability of PsyCrisisBench across linguistic and cultural contexts, a cross-lingual experiment was conducted. The Chinese transcripts were translated into English using Qwen3-max, with instructions to perform culturally-adapted translation that converts Chinese cultural contexts, emotional expressions, and social relationships into equivalent English expressions rather than literal word-by-word translation. The translated transcripts were then evaluated using QwQ-32B under both zero-shot and static few-shot conditions.

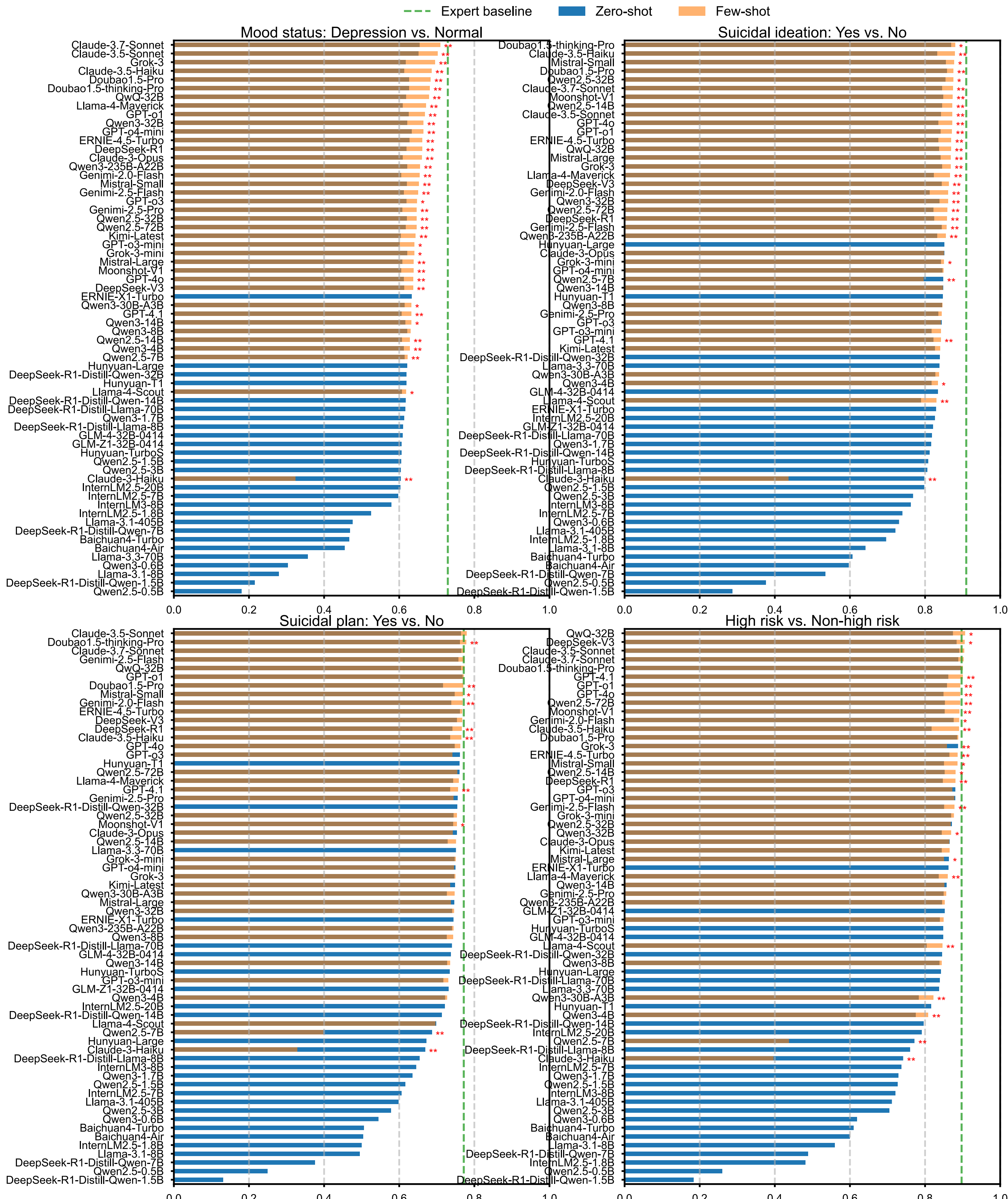

Fig. 3. Comparison of F1-Scores of LLMs on the Four PsyCrisisBench Tasks. This figure presents the F1-scores of 64 LLMs on four tasks: mood status recognition, suicidal ideation detection, suicide plan identification, and risk assessment. All models were evaluated under a zero-shot setting (blue bars). Among these, 38 models supporting long context windows were additionally evaluated under a static few-shot (4-shot) setting (orange bars). Asterisks indicate statistically significant differences between few-shot and zero-shot performance for the same model (* $p < 0.05$, ** $p < 0.01$, Welch's t-test). The green dashed line in each panel represents the human expert baseline, established by two trained hotline operators. The four examples used in the few-shot evaluation were carefully selected from an independent auxiliary dataset. For brevity, the "-Instruct" suffix is omitted from open-source model names (e.g., "Qwen2.5-7B" refers to "Qwen2.5-7B-Instruct").

### D. Implementation Details

For model invocation, all closed-source models were accessed via their respective official API interfaces. Among open-source models, InternLM2.5 (1.8B, 7B, 20B), InternLM3-8B, Qwen2.5 (0.5B~72B), Qwen3(0.6B~235B), QwQ-32B models, and all AWQ quantized models had their official Hugging Face weights downloaded and were deployed locally. Other open-source models were accessed via their corresponding vendors' online platforms.

Inference parameters (temperature, top_p, top_k) were configured following official recommendations from technical reports and documentation. Specifically, Mistral models used temperature=0.7; DeepSeek series used temperature=0.6; Qwen2.5 series used temperature=0.6 and top_p=0.9; Qwen3 series used temperature=0.6, top_p=0.95, and top_k=20 in non-reasoning mode, and temperature=0.7, top_p=0.8, and top_k=20 in reasoning mode; Moonshot models used temperature=0.3. Models without explicit recommendations used platform default parameters.

All local experiments were run on a machine equipped with an Intel Xeon Gold 6330 CPU and two NVIDIA A800 GPUs, using PyTorch and Hugging Face Transformers frameworks for model fine-tuning and inference deployment.

## IV. Results

### A. Overall Benchmark Performance

We systematically evaluated 64 LLMs on the PsyCrisisBench across four key psychological crisis assessment tasks: mood status recognition, suicidal ideation detection, suicidal plan identification, and risk assessment. Fig. 3 presents the detailed F1-score distributions for all models under both zero-shot and static few-shot condition, with the human expert baseline indicated by a green dashed line.

Collectively, the LLMs demonstrated robust performance in identifying suicidal ideation and assessing high-risk status, with multiple models achieving F1-scores exceeding 0.85, indicating substantial discriminative capability for these tasks. In contrast, mood status recognition proved more challenging, with most models yielding F1-scores between 0.6 and 0.7, highlighting the increased complexity of this task for current LLM reasoning.

Specifically, considering both zero-shot and static few-shot paradigms, the highest F1-scores were 0.709 [0.686, 0.731] for mood status recognition (Claude-3.7-Sonnet), 0.880 [0.865, 0.895] for suicidal ideation detection (Doubao1.5-thinking-Pro), 0.779 [0.756, 0.802] for suicidal plan identification (Claude-3.5-Sonnet), and 0.907 [0.891, 0.921] for risk assessment (QwQ-32B).

Compared to the human expert baseline, LLMs exhibited task-dependent performance patterns. Human experts achieved F1-scores of 0.729 [0.688, 0.767] for mood status recognition and 0.910 [0.886, 0.932] for suicidal ideation detection, outperforming the best LLMs on these two tasks. In contrast, the top-performing LLMs achieved comparable or slightly higher F1-scores than human experts on suicidal plan identification (LLM: 0.779 [0.756, 0.802] vs. Expert: 0.772 [0.725, 0.815]) and risk assessment (LLM: 0.907 [0.891, 0.921] vs. Expert: 0.897 [0.870, 0.923]). These results indicate that state-of-the-art LLMs have reached performance levels broadly comparable to trained human operators in text-based crisis assessment.

Several models, including DeepSeek-R1-Distill-Qwen-1.5B, Qwen2.5-0.5B, and Llama-3.1-8B, exhibited performance below random chance across tasks. This was primarily attributed to their failure to adhere to the mandated JSON format for structured output, leading to their responses being classified as incorrect under the evaluation protocol, irrespective of the substantive accuracy of their assessments.

### B. Zero-Shot vs Static Few-Shot Prompting Comparison

To evaluate the influence of prompting strategies, we compared the performance of 38 models under zero-shot and static few-shot (4-shot) conditions.

Overall, static few-shot prompting generally improved model F1-scores across tasks (Fig. 3). For mood status recognition, the top zero-shot F1-score of 0.654 [0.630, 0.677] (Claude-3.7-Sonnet) increased to 0.709 [0.686, 0.731] (Claude-3.7-Sonnet) with few-shot ($\Delta$F1 = 0.055). For suicidal ideation detection, the top F1-score rose from 0.869 [0.853, 0.884] to 0.880 [0.865, 0.895] (Doubao1.5-thinking-Pro) ($\Delta$F1 = 0.011). For suicidal plan identification, the top F1-score of 0.768 [0.744, 0.792] (GPT-o1) improved to 0.779 [0.756, 0.802] (Claude-3.5-Sonnet) ($\Delta$F1 = 0.011). For risk assessment, the top F1-score increased from 0.896 [0.880, 0.912] (Doubao1.5-thinking-Pro) to 0.907 [0.891, 0.921] (QwQ-32B) ($\Delta$F1 = 0.011). Mood status recognition exhibited the largest improvement magnitude.

For mood status recognition, 36 out of 38 models showed statistically significant ($p < 0.05$) improvements with few-shot prompting. The number of models demonstrating significant gains was 27 for suicidal ideation detection, 8 for suicidal plan identification, and 19 for risk assessment.

### C. Open-Source vs Closed-Source Models Comparison

We compared the performance of open-source ($n$=38) and closed-source ($n$=26) LLMs on the PsyCrisisBench. Fig. 4 illustrates the detailed F1-score distributions for both model categories.

Overall, open-source and closed-source models demonstrated comparable performance on most tasks, with leading open-source models matching or, in some instances, exceeding the capabilities of their closed-source counterparts. In suicidal ideation detection, the top-performing open-source model, Qwen3-32B (F1: 0.878 [0.863, 0.893]), achieved performance comparable to the leading closed-source model, Doubao1.5-thinking-Pro (F1: 0.880 [0.865, 0.895]). For suicidal plan identification, the best open-source model, QwQ-32B (F1: 0.773 [0.749, 0.797]), closely approached the performance of the top closed-source model, Claude-3.5-Sonnet (F1: 0.779 [0.756, 0.802]). In risk assessment, the leading open-source model, QwQ-32B (F1: 0.907 [0.891, 0.921]), surpassed the best-performing closed-source model, Claude-3.5-Sonnet (F1: 0.904 [0.890, 0.919]). However, closed-source models exhibited a discernible advantage in mood status recognition, where the top-performing closed-source model, Claude-3.7-Sonnet (F1: 0.709 [0.686, 0.731]), outperformed the best open-source model, QwQ-32B (F1: 0.679 [0.655, 0.702]).

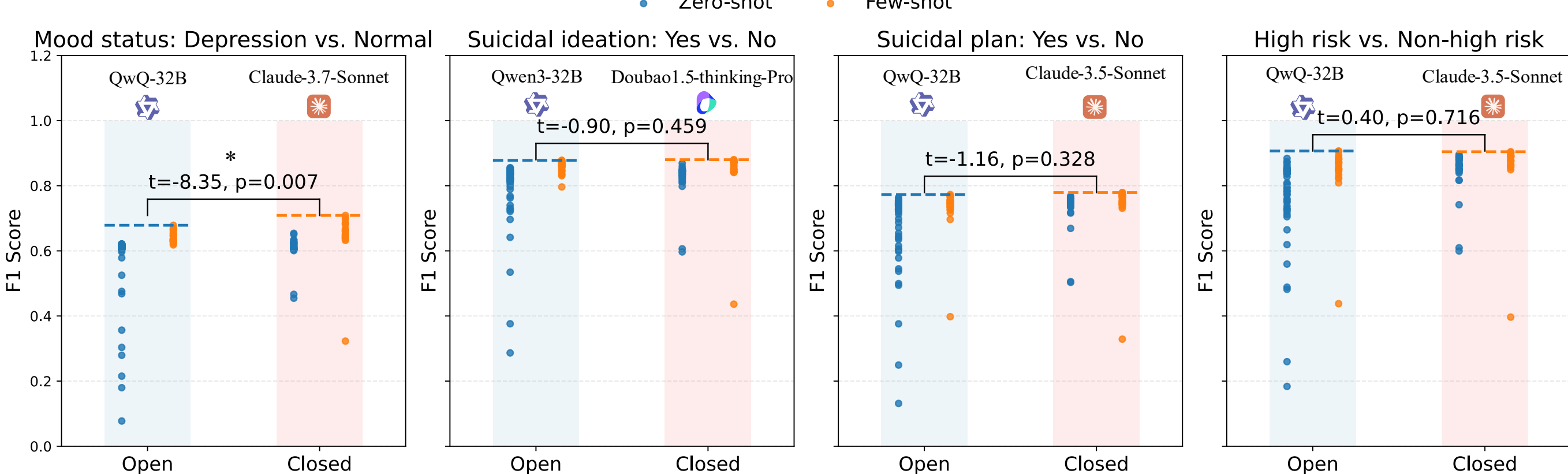

Fig. 4. F1-Score Performance Comparison of Open-Source and Closed-Source LLMs on PsyCrisisBench. This figure compares the F1-scores of open-source models (blue shaded area) and closed-source models (orange shaded area) across the four evaluation tasks. Scatter points represent the performance of individual models. In each task, the top-performing open-source and closed-source models are highlighted, and the statistical significance of their F1-score differences is examined using Welch's t-test. The t-scores and *p*-values are annotated in the figure.

Welch's t-test analysis revealed that in mood status recognition, the top-performing closed-source model (Claude-3.7-Sonnet) was significantly superior to the best open-source model (QwQ-32B) ($t$ = -8.35, $p$ = 0.007). Conversely, for suicidal ideation detection, suicidal plan identification, and risk assessment, the performance differences between the top-performing open-source models (Qwen3-32B for suicidal ideation, QwQ-32B for suicidal plan and high risk) and the top-performing closed-source models (Doubao1.5-thinking-Pro for suicidal ideation, Claude-3.5-Sonnet for suicidal plan identification and risk assessment) were not statistically significant (suicidal ideation: $t$ = -0.90, $p$ = 0.459; suicidal plan: $t$ = -1.16, $p$ = 0.328; high risk: $t$ = 0.40, $p$ = 0.716). These findings suggest that while closed-source models maintain an edge in tasks requiring nuanced affective understanding, state-of-the-art open-source LLMs can achieve comparable performance in tasks characterized by more explicit structural information.

## D. Reasoning vs Non-Reasoning Comparison

We further compared the performance of reasoning models against non-reasoning models on PsyCrisisBench (Fig. 5(a)). The categorization of models into reasoning and non-reasoning types is described in Section III-A. Across all four evaluation tasks, the F1-score differences between the top-performing reasoning and non-reasoning models were not statistically significant (mood status: $t$ = -2.00, $p$ = 0.165; suicidal ideation: $t$ = -0.42, $p$ = 0.706; suicidal plan: $t$ = 0.17, $p$ = 0.879; high risk: $t$ = -0.22, $p$ = 0.847). This suggests that, within the current evaluation framework, specialized reasoning architectures did not confer a distinct advantage for these crisis assessment tasks.

Additionally, we examined the performance of the Qwen3 series, which feature hybrid reasoning capabilities, by comparing their performance with reasoning mode explicitly activated versus non-activated (Fig. 5(b)). For mood status recognition, Qwen3 models in reasoning mode generally achieved significantly higher F1-scores than in non-reasoning mode, indicating that tasks demanding deeper semantic understanding may benefit from more sophisticated internal processing. However, for the other three tasks, performance differences between modes were inconsistent, with some models performing better in non-reasoning mode or showing no significant difference. This suggests that direct pattern matching may be sufficient for these tasks, and additional reasoning steps could introduce unnecessary complexity. Furthermore, reasoning mode was associated with substantially longer outputs: Qwen3 series models produced an average of 739.8 ± 454 words in reasoning mode compared to 165.03 ± 218.76 words in non-reasoning mode, a nearly 4.5-fold increase.

## E. Scaling Effects inside Model Families

We investigated the relationship between model parameter size and performance across the four tasks by examining models of varying scales within the same family, as illustrated in Fig. 6.

A general trend indicates that F1-scores typically improve with increasing parameter count across most tasks and model series. This trend was particularly clear and consistent within the InternLM2.5 series: for instance, in risk assessment, the F1-score steadily rose from 0.482 [0.452, 0.512] for the 1.8B model to 0.792 [0.768, 0.814] for the 20B model. Similar scaling patterns were consistently observed across other model families.

However, beyond a certain parameter scale, marginal performance gains often diminished. The Qwen2.5 series exemplifies this: a substantial leap in performance was observed when scaling from 0.5B to 1.5B parameters (e.g., mood status recognition F1 increased from 0.180 [0.154, 0.208] to 0.605 [0.580, 0.629]), but subsequent gains became inconsistent, with performance plateauing or even declining at larger scales (e.g., suicidal ideation F1 peaked at 0.854 [0.838, 0.870] for 32B before declining to 0.822 [0.805, 0.839] for 72B). Similar saturation patterns were observed across other model families.

Task characteristics also influenced scaling effects. Mood

(a)

(b)

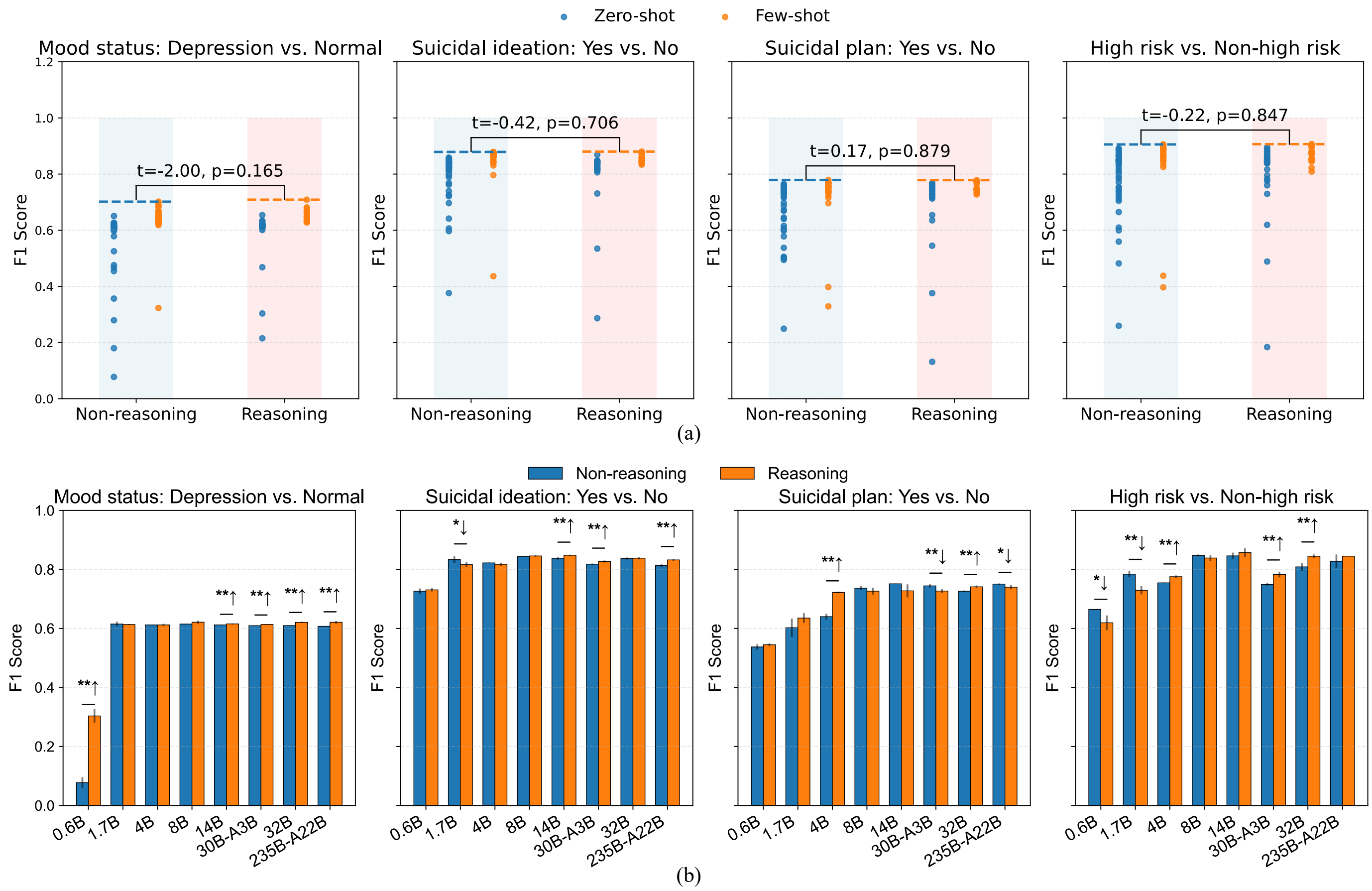

Fig. 5. F1-Score Performance Comparison of Reasoning and Non-Reasoning LLMs on PsyCrisisBench. (a) This figure compares the F1-scores of non-reasoning models (blue shaded area) and reasoning models (orange shaded area) across the four evaluation tasks. Scatter points represent the performance of individual models. In each task, the top-performing non-reasoning and reasoning models are highlighted, and the statistical significance of their F1-score differences is examined using Welch's t-test. The t-scores and $p$-values are annotated in the figure. (b) This panel shows the F1-scores of the Qwen3 series of hybrid-reasoning models in non-reasoning mode (blue bars) and reasoning mode (orange bars) across the four tasks. Asterisks indicate statistically significant differences in performance between the two modes for the same model (* $p < 0.05$, ** $p < 0.01$, Welch's t-test).

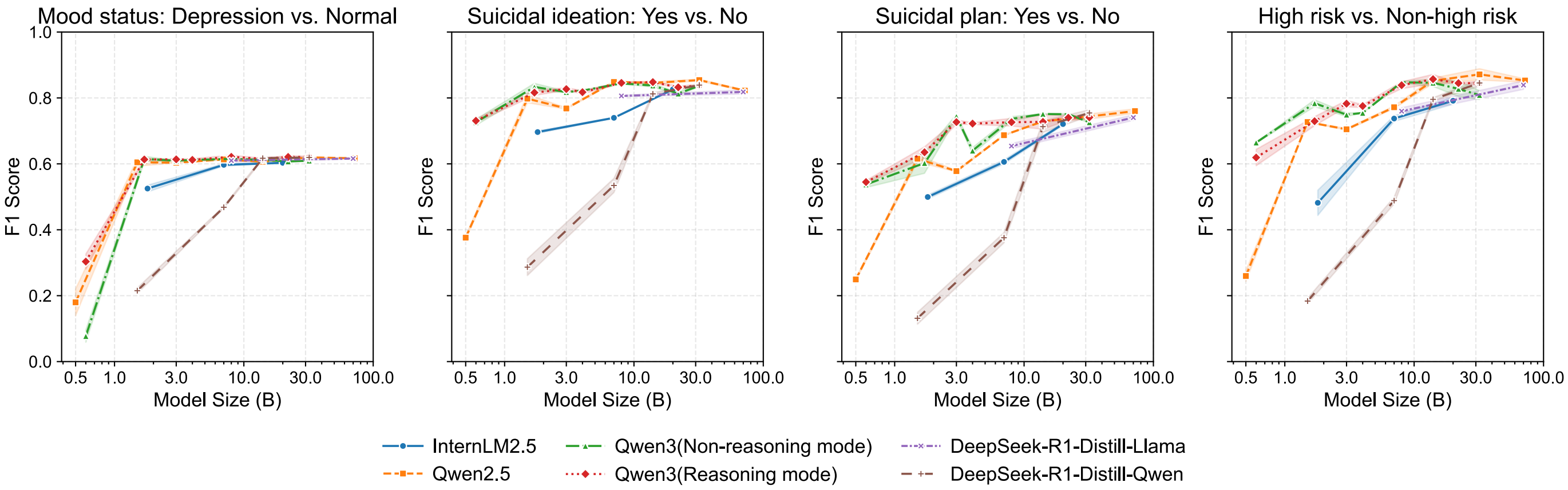

Fig. 6. Scaling effects across LLM families. F1-scores of models with different parameter sizes (log-scaled, in billions) are shown across the four tasks. Lines connect models from the same series to highlight trends in performance relative to scale.

status recognition showed earlier performance saturation than other tasks, while suicidal ideation detection, suicidal plan identification, and risk assessment continued to benefit from larger parameter counts, albeit eventually encountering diminishing returns.

## F. Dynamic Few-Shot and Fine-Tuning Analysis

To further investigate the effects of few-shot example quantity and domain-specific fine-tuning, we conducted ablation studies (Fig. 7).

In dynamic few-shot experiments using Qwen2.5-7B-

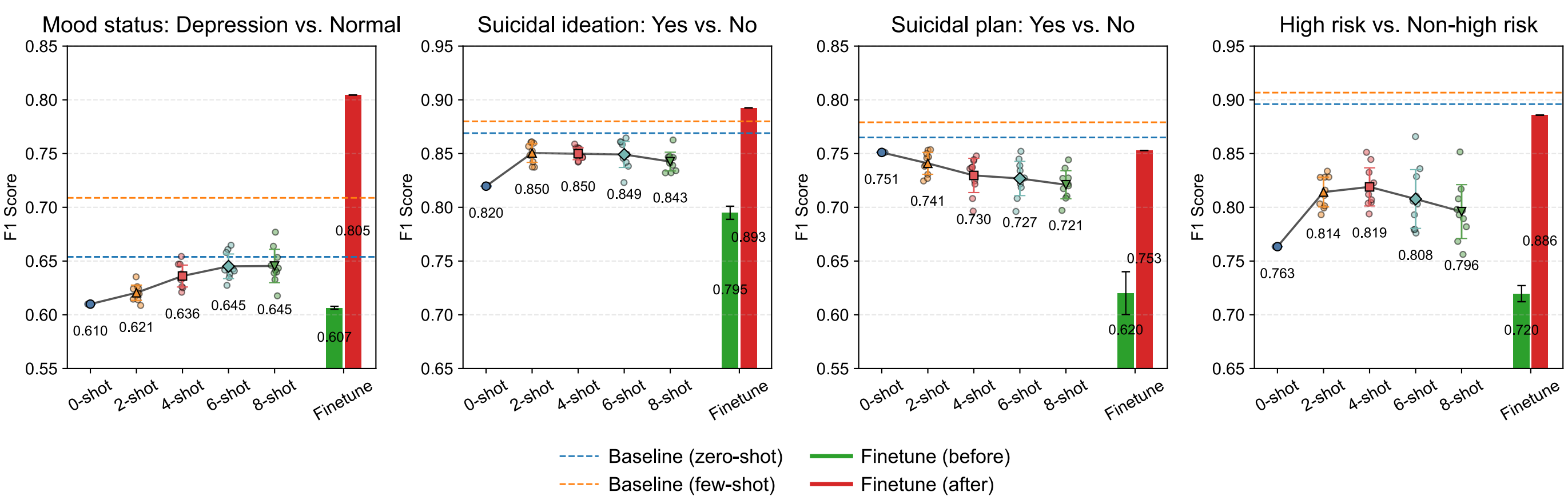

Fig. 7. Dynamic few-shot and fine-tuning performance of Qwen models. The figure presents F1-scores of Qwen2.5-7B across zero-shot and dynamic few-shot settings (2, 4, 6, 8-shot, repeated 10 times with random samples). Dark points indicate the mean, with error bars showing standard deviation. The second panel shows Qwen2.5-1.5B performance before and after fine-tuning on 520 cases from 2022 auxiliary dataset. Dashed lines indicate the best-performing zero-shot (blue) and static 4-shot (orange) baselines across all models.

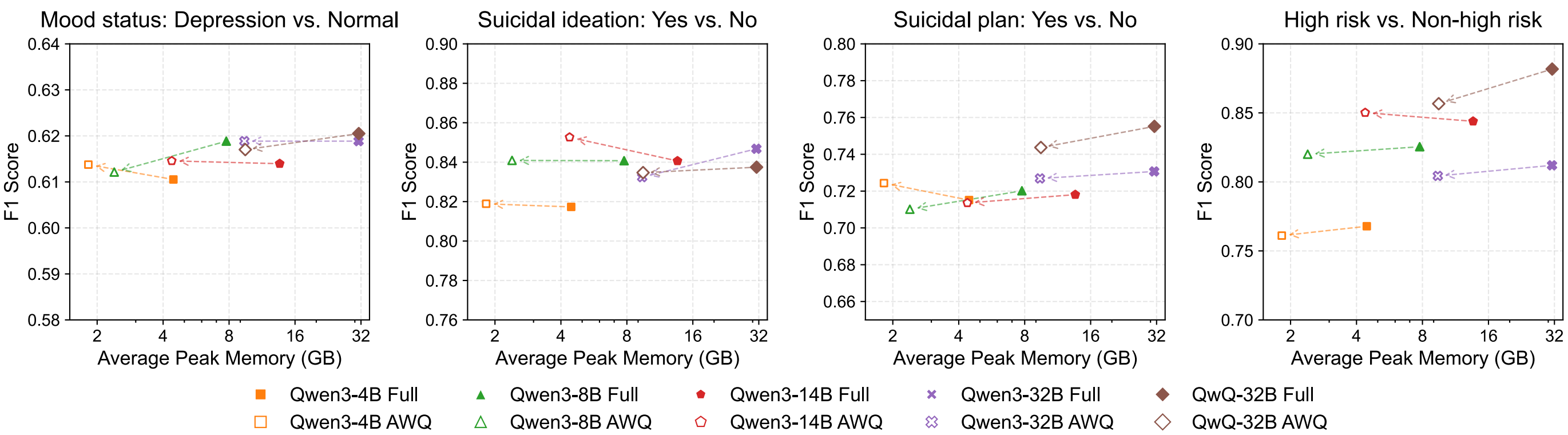

Fig. 8. Performance and memory comparison of full-precision vs. quantized models. F1-scores and peak GPU memory usage are compared between full-precision and AWQ-quantized versions of selected open-source models. Solid dots indicate full-precision models; hollow dots represent their quantized counterparts. Dashed arrows trace memory reduction and performance shift within the same model series.

Instruct-1M, model performance was assessed under zero-shot and 2, 4, 6, 8-shot conditions, with each condition repeated 10 times using different randomly sampled examples. For mood status recognition, performance steadily increased with more examples: from 0.610 [0.568, 0.649] at zero-shot to 0.621 [0.608, 0.633] at 2-shot, 0.636 [0.623, 0.649] at 4-shot, and plateauing at 0.645 [0.632, 0.658] for 6 and 8-shot, suggesting that more contextual examples aid model calibration. Suicidal ideation detection improved from 0.820 [0.789, 0.849] at zero-shot to 0.850 [0.841, 0.859] at 2-shot, then slightly declined at higher shots (0.843 [0.833, 0.852] at 8-shot), though all few-shot conditions outperformed zero-shot. In contrast, suicidal plan identification showed progressive decline: from 0.751 [0.707, 0.793] at zero-shot to 0.721 [0.707, 0.735] at 8-shot, suggesting high sensitivity to example quality where suboptimal random samples may introduce noise. Risk assessment peaked at 4-shot (0.819 [0.807, 0.828]) compared to zero-shot (0.763 [0.727, 0.797]), with slight decline at 6 and 8 shots but still substantially better than zero-shot.

Fine-tuning Qwen2.5-1.5B using the 2022 auxiliary dataset yielded substantial improvements: mood status recognition from 0.607 [0.582, 0.631] to 0.805 [0.783, 0.825]; suicidal ideation detection from 0.795 [0.776, 0.813] to 0.893 [0.877, 0.907]; suicidal plan identification from 0.620 [0.589, 0.649] to 0.753 [0.726, 0.778]; and risk assessment from 0.720 [0.697, 0.741] to 0.886 [0.869, 0.901]. Notably, the fine-tuned model, despite its small parameter size, surpassed the best zero-shot and static few-shot performance for mood status recognition and suicidal ideation detection, and achieved highly competitive results for the other two tasks.

## G. Quantization Trade-Offs

To evaluate the trade-off between computational efficiency and model performance, we compared full-precision models with their AWQ-quantized counterparts under zero-shot evaluation (Fig. 8). The comparison included Qwen3 series (4B/8B/14B/32B) and QwQ-32B.

AWQ quantization consistently achieved substantial memory reduction (60-70%) across all models. For example, average peak GPU memory decreased from 31.3 GB to 9.5 GB for QwQ-32B and from 31.2 GB to 9.4 GB for Qwen3-32B, with similar reduction ratios observed for smaller models.

Crucially, these memory savings were achieved with minimal performance degradation. For QwQ-32B, the largest performance gap was observed in risk assessment (Full: 0.882 [0.853, 0.908] vs. AWQ: 0.857 [0.825, 0.886]), while other tasks showed negligible differences. The Qwen3 series demonstrated even greater robustness to quantization, with Qwen3-14B showing comparable or slightly improved performance after quantization. These results suggest that AWQ quantization offers a practical solution for deploying LLMs in resource-constrained settings without substantially compromising crisis assessment accuracy.

### H. Cross-Lingual Evaluation on English Translations

To assess the generalizability of PsyCrisisBench across linguistic contexts, we evaluated QwQ-32B on English translations of the Chinese transcripts under both zero-shot and few-shot conditions (Table II).

Overall, QwQ-32B maintained comparable performance on English translations across most tasks. Mood status recognition showed virtually unchanged zero-shot performance with slight few-shot decrease. Interestingly, suicidal ideation detection showed marginally higher performance on English translations under zero-shot conditions. The largest performance gaps were observed in suicidal plan identification ($\Delta F1 \approx 0.02$ decrease), while risk assessment showed consistent but modest decreases across both conditions.

TABLE II

CROSS-LINGUAL PERFORMANCE COMPARISON OF QWQ-32B.

| Task | Setting | Chinese (Original) | English (Translated) |
|---|---|---|---|
| Mood status | Zero-shot | 0.618 [0.594, 0.641] | 0.619 [0.595, 0.642] |
| | Few-shot | 0.679 [0.655, 0.702] | 0.665 [0.641, 0.688] |
| Suicidal ideation | Zero-shot | 0.837 [0.819, 0.853] | 0.851 [0.834, 0.867] |
| | Few-shot | 0.869 [0.853, 0.884] | 0.863 [0.846, 0.878] |
| Suicidal plan | Zero-shot | 0.764 [0.739, 0.788] | 0.741 [0.716, 0.765] |
| | Few-shot | 0.773 [0.749, 0.797] | 0.753 [0.728, 0.778] |
| Risk assessment | Zero-shot | 0.874 [0.857, 0.890] | 0.858 [0.840, 0.874] |
| | Few-shot | 0.907 [0.891, 0.921] | 0.891 [0.875, 0.906] |

## V. DISCUSSION

This study establishes PsyCrisisBench as a novel and clinically grounded benchmark for evaluating LLMs in crisis triage, specifically within the domain of psychological support hotlines. Our results demonstrate that state-of-the-art LLMs can achieve performance broadly comparable to trained human operators in text-based crisis assessment, with strong capabilities in suicide-related detection and risk assessment. However, we also identified key limitations—particularly in mood state inference—that highlight persistent challenges in affective reasoning and real-world clinical alignment.

LLMs, particularly from the Claude and Qwen families, demonstrated robust zero-shot performance on structured tasks involving suicidal ideation detection, suicide plan identification, and risk assessment. These tasks appear to benefit from the explicit linguistic markers in psychological hotline transcripts, facilitating generalization without domain-specific adaptation. Specifically, the Claude family models (Claude-3.7-Sonnet, Claude-3.5-Sonnet, Claude-3.5-Haiku) and Doubao1.5-thinking-Pro consistently ranked among the top performers across all tasks, with F1-scores approaching or exceeding 0.85 in some cases, highlighting the potential of closed-source, instruction-tuned models for high-stakes classification tasks. However, performance on mood status recognition was universally lower, with no model surpassing an F1-score of 0.71.

Two factors likely contribute to this underperformance. First, emotional tone and mood cues are often embedded in prosodic and paralinguistic features [35] that are lost in transcription, weakening the signal available for text-based models. Second, the definition of "depression" in PsyCrisisBench adheres to stricter psychiatric diagnostic criteria [36], whereas general-purpose LLMs may default to colloquial interpretations of the term—conflating transient sadness with clinical depression. This mismatch partially explains why static few-shot and fine-tuned paradigms yield particularly large gains in mood status recognition: the in-context examples and labeled training instances help realign the model's interpretation of depressive language with clinically valid thresholds. Notably, dynamic few-shot prompting showed continued improvement as more examples were added, suggesting that sample-rich prompts serve as effective anchors for clarifying ambiguous labels.

The comparison with human expert baseline provides further insight into these task-dependent performance patterns. Human operators outperformed the best LLMs on mood status recognition and suicidal ideation detection, while LLMs achieved comparable or slightly superior performance on suicidal plan identification and risk assessment. It is important to note that all evaluations were conducted on transcribed text containing recognition errors. These differences likely reflect the distinct cognitive strengths of humans and LLMs. Mood status recognition and suicidal ideation detection may require greater empathy and nuanced emotional interpretation—capabilities that experienced operators have developed through extensive clinical practice. In contrast, suicidal plan identification and risk assessment depend more heavily on detecting specific factual details scattered across lengthy transcripts, which typically span thousands to tens of thousands of words. LLMs' attention mechanisms enable precise localization of key textual cues and semantic reconstruction from noisy input, whereas human operators—who are not specifically trained to evaluate solely from text transcripts—may overlook critical details during extended reading. Overall, these findings suggest that under text-only conditions, trained operators and LLMs demonstrate broadly comparable performance, each with complementary strengths.

While prior studies in domains such as nephrology have shown that closed-source models like GPT-4 substantially outperform open models [37], our comparative analysis between open- and closed-source models highlights an encouraging trend: the strongest open models, particularly QwQ-32B, rivaled commercial systems such as Claude 3.7 in most tasks. No statistically significant difference was observed in suicidal ideation detection, suicide plan identification or risk assessment ($p$-values all > 0.3), suggesting that, with proper prompting, open models can match the reliability of their proprietary counterparts in structurally explicit classification settings. Nonetheless, closed-source models retained a distinct advantage in mood classification ($p = 0.007$), reinforcing the importance of high-quality alignment and emotional reasoning—features that remain more advanced in proprietary models trained on broader, richer human feedback datasets.

Few-shot prompting yielded task-specific gains, particularly in mood classification, where performance scaled linearly with the number of in-context examples. In contrast, gains plateaued or declined for other tasks, suggesting that linguistic salience governs the marginal utility of additional demonstrations.

Nonetheless, the selection of in-context examples proved highly influential. For certain tasks, such as suicidal planning, we observed a counterintuitive performance decline as the number of examples increased beyond four. This may stem from prompt window saturation or conflicts introduced by randomly sampled cases. Suicidal planning likely demands more nuanced context, making it particularly sensitive to the quality and consistency of examples. This hypothesis is supported by our targeted few-shot trials (Fig. 3), where carefully curated four-example prompts consistently outperformed zero-shot baselines. These findings underscore that quantity alone does not guarantee better performance—quality and relevance of examples are equally critical [38].

Fine-tuning, in contrast, delivered consistent and substantial improvements. These findings are consistent with recent evaluations in clinical oncology [39]. The Qwen-2.5-1.5B model, despite its smaller size, achieved performance that exceeded even dynamic few-shot prompting after task-specific training. For example, suicidal plan identification improved from 0.62 (zero-shot) to 0.75 (post-finetuning), indicating that task alignment, rather than scale, is the primary determinant of reliability in sensitive domains [40]. In mood classification and suicidal ideation detection, the fine-tuned 1.5B model even surpassed all zero-shot and few-shot baselines, including trillion-parameter commercial systems, further underscoring the value of targeted domain adaptation. This finding provides a practical pathway for healthcare institutions to develop cost-effective, privacy-preserving AI solutions by leveraging their own clinical data.

Scaling trends across LLM families demonstrated a common pattern [41]: performance improved with model size up to 13B parameters, beyond which gains plateaued. The Qwen3-reasoning series outperformed its non-reasoning counterparts, particularly in complex inference tasks, suggesting that architectural enhancements or alignment tuning for deductive reasoning may offer greater returns than parameter count alone. However, when comparing the best reasoning and non-reasoning models across all tasks, differences were not statistically significant, indicating that reasoning modules may not always translate into downstream advantages. Moreover, reasoning-enabled models often incur additional latency and computational cost due to longer output sequences.

Recent trends in LLM architecture highlight hybrid reasoning strategies capable of toggling between reasoning and non-reasoning modes depending on prompt structure or task difficulty. This is evident in models such as Gemini-2.5-Flash and Qwen3, where flexible decoding pathways enable more adaptive inference. In our evaluation, the reasoning variant of Qwen3 showed notable gains in mood classification, potentially due to enhanced ability to interpret ambiguous affective signals—reaffirming the need for robust emotional reasoning in mental health contexts. However, the 4.5-fold increase in output length associated with reasoning mode presents a practical trade-off: while more detailed reasoning processes may enhance decision transparency, whether such verbosity benefits clinical practice remains uncertain. Clinically meaningful explainability should align with established crisis intervention workflows rather than reflect model-specific reasoning patterns shaped by their training data; without such alignment, extended outputs may increase operator burden rather than support decision-making. Quantized models demonstrated minimal performance loss while significantly reducing inference cost, making them attractive for deployment in resource-limited settings where computational efficiency and privacy preservation are essential.

Our cross-lingual evaluation on English translations provides preliminary evidence that PsyCrisisBench's task structure possesses a degree of language-agnostic validity. However, this translation-based approach represents a simplified approximation of true cross-cultural evaluation and cannot fully capture the linguistic and cultural nuances inherent in authentic multilingual crisis communication.

### *Limitations and Future Directions*

This study has several limitations. First, PsyCrisisBench is derived from a single Mandarin-speaking hotline over one year, which may limit generalizability to other languages, populations, and intervention settings. Our cross-lingual evaluation on English translations provides only a simplified approximation of cross-cultural transfer; future work should evaluate the framework on authentic multilingual datasets. Second, the use of binary labels for mood status and suicide-related constructs, although grounded in a structured clinical questionnaire, necessarily compresses dynamic and continuous psychological states into discrete categories. Third, all assessments were based on automatic speech recognition transcripts: recognition errors introduce noise, and crucial nonverbal cues such as prosody, pace, and hesitation are lost, likely imposing a performance ceiling for both humans and LLMs under text-only conditions. Fourth, the strict requirement for JSON-formatted output means that models with otherwise reasonable clinical judgments can be scored poorly if they violate formatting rules, implying that our benchmark may underestimate some models' underlying reasoning capabilities. Finally, the current evaluation is static and single-turn, and therefore does not capture longitudinal changes or multi-turn conversational strategies that are central to real-world crisis intervention.

Looking forward, future research should prioritize multimodal integration incorporating prosodic and paralinguistic features, longitudinal evaluation frameworks that capture the dynamic nature of crisis intervention, and rigorous cross-cultural validation across diverse linguistic and clinical contexts. Additionally, exploring guided reasoning approaches that integrate model inference with clinical protocols could enhance both transparency and practical utility. As the performance gap between open- and closed-source models continues to narrow, the field should also prioritize interpretability and human-AI collaboration to develop safe, context-aware, and ethically aligned crisis response systems.

## VI. Conclusion

Our benchmark demonstrates that advanced LLMs can achieve performance comparable to trained human operators on structured crisis assessment tasks, while revealing persistent challenges in affective reasoning that warrant targeted enhancement. PsyCrisisBench offers a replicable, real-world foundation for ongoing model development and evaluation in digital mental health. In particular, our results suggest a clear performance hierarchy—fine-tuning > few-shot > zero-shot—highlighting the importance of domain adaptation in emotionally nuanced tasks. For clinical deployment, especially in resource-constrained settings, a practical and effective strategy involves fine-tuning smaller open-source models using historical conversation data, followed by model quantization to optimize inference efficiency. This approach not only reduces computational costs and preserves privacy, but also delivers performance comparable to larger, proprietary systems. As the performance gap between open- and closed-source LLMs continues to narrow, transparent and adaptable systems with strong emotional alignment to clinical constructs will become increasingly viable for real-world deployment.

## Appendix A

### Annotation Protocol for Hotline Crisis Assessment

All annotations in the PsyCrisisBench dataset were derived from a standardized clinical assessment conducted immediately after each call. Trained hotline operators systematically evaluated callers using the Depression and Suicide Risk Assessment Questionnaire (Table III).

The Mood Status label was determined by the severity of depressive symptoms based on the questionnaire responses. A caller was classified as “Normal” (no depression) if both Item 1 and Item 2 were answered “No” and no suicidal ideation was reported. The “Depression” label encompassed three severity levels: (1) Mild depression required an affirmative answer to either Item 1 or Item 2, at least four affirmative answers among Items 1–9, and an affirmative answer to either Item 10 or Item 11. (2) Moderate depression required an affirmative answer to either Item 1 or Item 2, five to seven affirmative answers among Items 1–9, and an affirmative answer to either Item 10 or Item 11. (3) Major depression required an affirmative answer to either Item 1 or Item 2 with a duration of at least one week, seven or more affirmative answers among Items 1–9, and an affirmative answer to either Item 10 or Item 11.

Suicidal Ideation was labeled “Yes” if the caller answered “Yes” to Item 12, and “No” otherwise.

Suicidal Plan was labeled “Yes” if the caller answered “Yes” to Item 13, and “No” otherwise.

The Risk Assessment label followed a two-criterion rule. A caller was classified as “High-risk” if they reported both suicidal ideation (Item 12 “Yes”) and a specific suicide plan (Item 13 “Yes”). Alternatively, a caller with suicidal ideation (Item 12 “Yes”) but no specific plan (Item 13 “No”) was classified as “High-risk” if their cumulative risk-factor score reached five or more points. Risk-factor scoring was as follows: one point each for a history of suicide attempts (Item 14 “Yes”), exposure to suicide in close contacts (Item 15 “Yes”), and a hopefulness score ≤ 50 % (Item 16 ≤ 50). One additional point was assigned if either serious physical illness/major life events (Item 17) or substance abuse (Item 18) was reported as present. Furthermore, one point was added for moderate depression and two points for severe depression. Callers not meeting either criterion were labeled “Non-high risk”.

TABLE III

Depression and Suicide Risk Assessment Questionnaire.

| Item | Question | Answer |
|---|---|---|
| 1 | Have you felt down or depressed recently? (sad face, sighing, tears alone) | Yes (duration) / No |
| 2 | Have you recently lost interest in things you normally enjoy? | Yes (duration) / No |
| 3 | Have you recently felt worthless, a useless person or a failure? | Yes / No |
| 4 | Do you blame yourself or have regrets? | Yes / No |
| 5 | Has your weight changed recently? | Yes (increase, decrease) / No |
| | Any recent changes in your appetite? | Yes (increase, decrease) / No |
| 6 | Has your sleep changed recently? | Yes (increase, decrease) / No |
| 7 | Have you had any physical complaints recently? (Dizziness, fatigue, listlessness or lack of energy) | Yes / No |
| 8 | Have you recently had trouble concentrating, thinking flexibly, or being stupid? | Yes / No |
| 9 | Did others see that you were different than usual (expression, speech, behavior, reaction)? | Yes / No |
| 10 | Has the above problem affected your work, study, life, socialization, etc.? | Yes / No |
| 11 | Do you feel distressed when these problems arise? | Yes / No |
| 12 | Many people have thought about death when they are in the above situations; have you ever thought about harming yourself? | Yes / No |
| 13 | Do you have a specific plan for self-injury or suicide? | Yes (describe: ) / No |
| | What is the intended time frame for this program? | Yes (describe: ) / No |
| | Do you really want to die? | Yes / No |
| 14 | Have you ever committed suicide or intentionally harmed yourself before? | Yes (describe: ) / No |
| | How many suicidal behaviors were there? | Times |
| | When did this last happen? | Year:Month:Day |
| | Did you really want to die at that time? | Yes / No |
| 15 | Have any of your relatives, friends, co-workers, or acquaintances ever self-injured or committed suicide? | Yes / No |
| 16 | If “0” means no hope and “100” means the most hope, how hopeful are you about your future life? | ≤50 / >50 |
| 17 | In the past month, have you had any serious physical illnesses, major life events? | Yes / No |
| 18 | In the last month, have you consumed excessive alcohol, abused sleeping pills, narcotics or stimulants? | Yes / No |
| 19 | What are the main causes of your suicidal thoughts? | |
| 20 | What problem do you hope to solve or achieve through suicide? | |

## Appendix B

## Auxiliary Dataset from Year 2022 and Few-Shot Sample Selection

To prevent data leakage between prompting examples and evaluation data, an independent auxiliary dataset was compiled from 520 calls selected from 16,573 calls made between January 1, 2022, and December 31, 2022. This dataset was processed using the same transcription pipeline and annotation criteria as PsyCrisisBench (see Section II and Appendix A). The demographic, call characteristics, and class distributions of this auxiliary dataset are presented in Fig. 9. The average audio duration was 35.7 ± 22.9 minutes, and the average transcript word count was 6,884 ± 4,929 words per call.

For static few-shot evaluation, four representative examples were selected from this auxiliary dataset based on two criteria. First, examples were chosen to balance positive and negative instances for the classification tasks, ensuring that models receive exposure to both crisis and non-crisis cases during few-shot prompting. Second, transcript lengths were controlled to approximate the median word count of PsyCrisisBench samples (5,101 words), avoiding extreme lengths that could either lack representativeness or exceed context window limitations. The characteristics of the four selected examples are presented in Table IV.

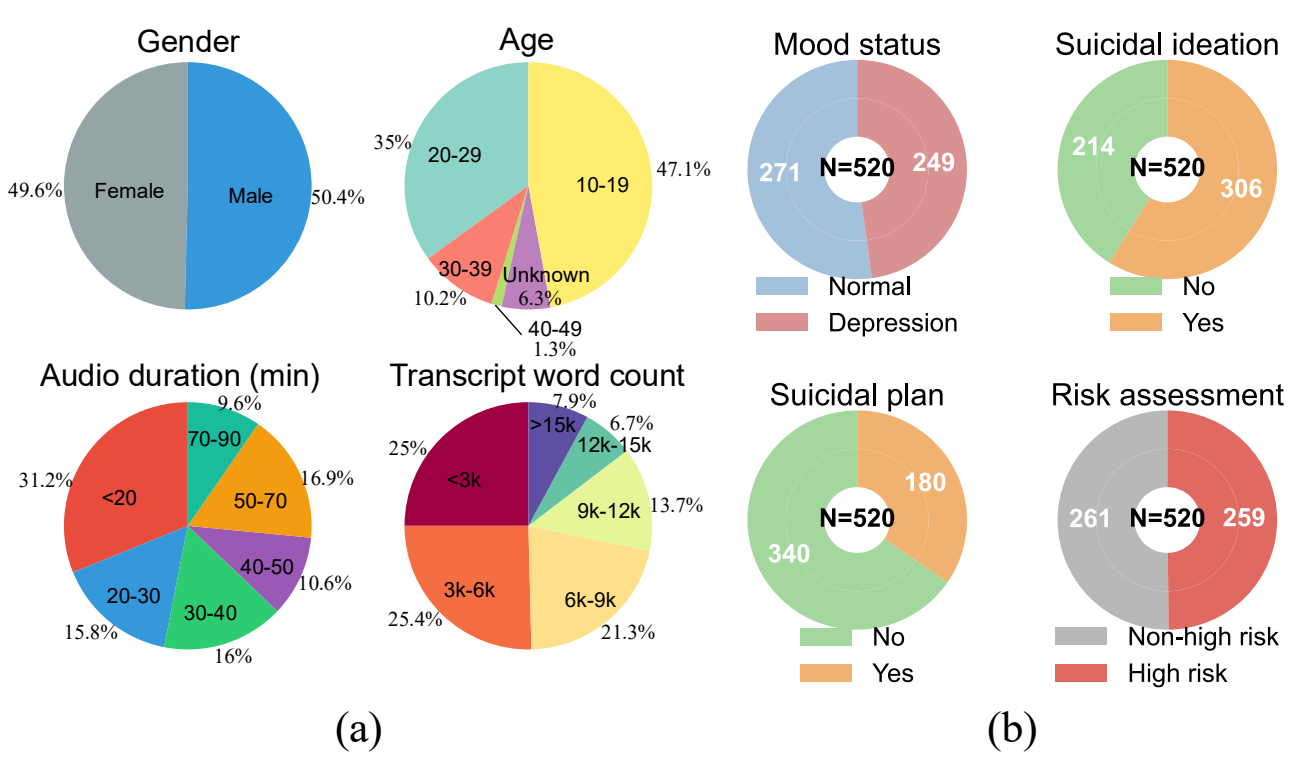

Fig. 9. Statistics of the 2022 auxiliary dataset used for few-shot prompting and fine-tuning experiments (*N*=520). (a) Pie charts showing the demographic and call characteristics, including distribution of gender, age, audio duration in minutes, and transcript word count. (b) Class distribution for the four binary classification tasks: mood status (Depression: 249, Normal: 271), suicidal ideation (Yes: 306, No: 214), suicide plan (Yes: 180, No: 340), and risk assessment (High risk: 259, Non-high risk: 261).

TABLE IV

Details of Four Examples Used in Static Few-Shot Evaluation

| Mood status | Suicidal ideation | Suicidal plan | Risk assessment | Transcript word count |
|---|---|---|---|---|
| Normal | No | No | Non-high risk | 3493 |
| Normal | No | No | Non-high risk | 4010 |
| Depression | Yes | Yes | High risk | 4929 |
| Depression | Yes | Yes | High risk | 6076 |